\documentclass[letterpaper, 10 pt, conference]{ieeeconf}  
\usepackage{amssymb}
\usepackage{amsmath}
\usepackage{graphicx} 
\usepackage{multirow}
\usepackage{subfig}
\usepackage{cite}
\usepackage[colorlinks=true,linkcolor=black,citecolor=blue,urlcolor=blue,pdftex]{hyperref}

\IEEEoverridecommandlockouts                              

\title{\LARGE \bf
Minimizing Acoustic Noise: Enhancing Quiet Locomotion for Quadruped Robots in Indoor Applications
}

\author{Zhanxiang Cao$^{1}$, Buqing Nie$^{1}$, Yang Zhang$^{2}$, and Yue Gao$^{3\dag}$
\thanks{*This work is supported by the National Natural Science Foundation of China (Grant No. 92248303 and No. 62373242), the Shanghai Municipal Science and Technology Major Project (Grant No. 2021SHZDZX0102), and the Fundamental Research Funds for the Central Universities.}
\thanks{$^{1}$Zhanxiang Cao and Buqing Nie are with Department of Computer Science and Engineering, Shanghai Jiao Tong University, Shanghai, P.R. China. Email: caozx1110@sjtu.edu.cn, niebuqing@sjtu.edu.cn}%
\thanks{$^{2}$Yang Zhang is with Department of Automation, Shanghai Jiao Tong University, Shanghai, P.R. China. Email: zhangyang-sjtu-2022@sjtu.edu.cn}%
\thanks{$^{3}$Yue Gao is with MoE Key Lab of Artificial Intelligence and AI Institute, Shanghai Jiao Tong University, Shanghai, P.R. China. Email: yuegao@sjtu.edu.cn}%
\thanks{$\dag$ Corresponding author.}%
}

\begin{document}

\maketitle
\thispagestyle{empty}
\pagestyle{empty}

\begin{abstract}
    Recent advancements in quadruped robot research have significantly improved their ability to traverse complex and unstructured outdoor environments. However, the issue of noise generated during locomotion is generally overlooked, which is critically important in noise-sensitive indoor environments, such as service and healthcare settings, where maintaining low noise levels is essential. This study aims to optimize the acoustic noise generated by quadruped robots during locomotion through the development of advanced motion control algorithms. To achieve this, we propose a novel approach that minimizes noise emissions by integrating optimized gait design with tailored control strategies. This method achieves an average noise reduction of approximately 8 dBA during movement, thereby enhancing the suitability of quadruped robots for deployment in noise-sensitive indoor environments. Experimental results demonstrate the effectiveness of this approach across various indoor settings, highlighting the potential of quadruped robots for quiet operation in noise-sensitive environments.
\end{abstract}

\section{INTRODUCTION}
Quadruped robots have garnered significant attention in recent years, particularly due to their versatility and capability to navigate complex terrains using Reinforcement Learning-based motion control~\cite{hwangbo2019learning, lee2020learning, miki2022learning, hoeller2024anymal, cheng2024extreme, kumar2021rma, margolis2023walk}. This adaptability makes them suitable for a variety of applications, including search and rescue~\cite{li2023fabrication}, exploration~\cite{miller2020mine}, and service industries~\cite{ma2020cloud}. Despite the remarkable success in outdoor and rugged tasks, applications of quadruped robots in indoor areas with noise-sensitivity are not fully researched, such as hospitals~\cite{de2021environmental}, offices~\cite{yadav2021sound}, and residential spaces~\cite{gilani2021study}—the demand for quiet operation becomes increasingly critical. In such scenarios, excessive noises generated during robot locomotion are unexpected or even disruptive, thus limiting the practical deployment of quadruped robots in such environments.

Although research in quadruped robotics has made impressive strides in motion control and terrain adaptability, the noise generated by these robots during operation are not taken into consideration in recent researches. As these robots are increasingly deployed in proximity to human users, especially in indoor spaces or environments with specific noise constraints, the noise they produce becomes a critical concern. Noise pollution may result in inconvenience, give rise to unexpected disturbances and, in certain circumstances, even impede task performance, especially in applications where stealth or subtlety is of utmost importance~\cite{jhanwar2016noise}. For example, the loud noise generated by a robot's footstep impact or motor functions not only causes discomfort to nearby humans, but also reduces the robot's effectiveness in scenarios that require quiet operation. Therefore, addressing this issue becomes critical to ensuring their seamless integration into human environments.

\subsection{Noise in Indoor Environments}
Noise pollution in indoor environments is not merely an inconvenience; it can have significant implications. In hospitals, for example, unwanted noise can interfere with patient recovery, increase stress levels for both patients and medical staff, and disrupt the overall healing atmosphere~\cite{de2021environmental}. In offices, excessive noise may reduce productivity, impair concentration, and negatively affect employee well-being~\cite{yadav2021sound}. Similarly, in residential areas, constant mechanical noise can cause discomfort and irritation to occupants, affecting the quiet and relaxing environments expected in these scenarios~\cite{gilani2021study}. Therefore, reducing the noise generated by quadruped robots in such environments is crucial for their widespread acceptance and utility.

\begin{figure}
    \centering
    \includegraphics[width=0.38\textwidth]{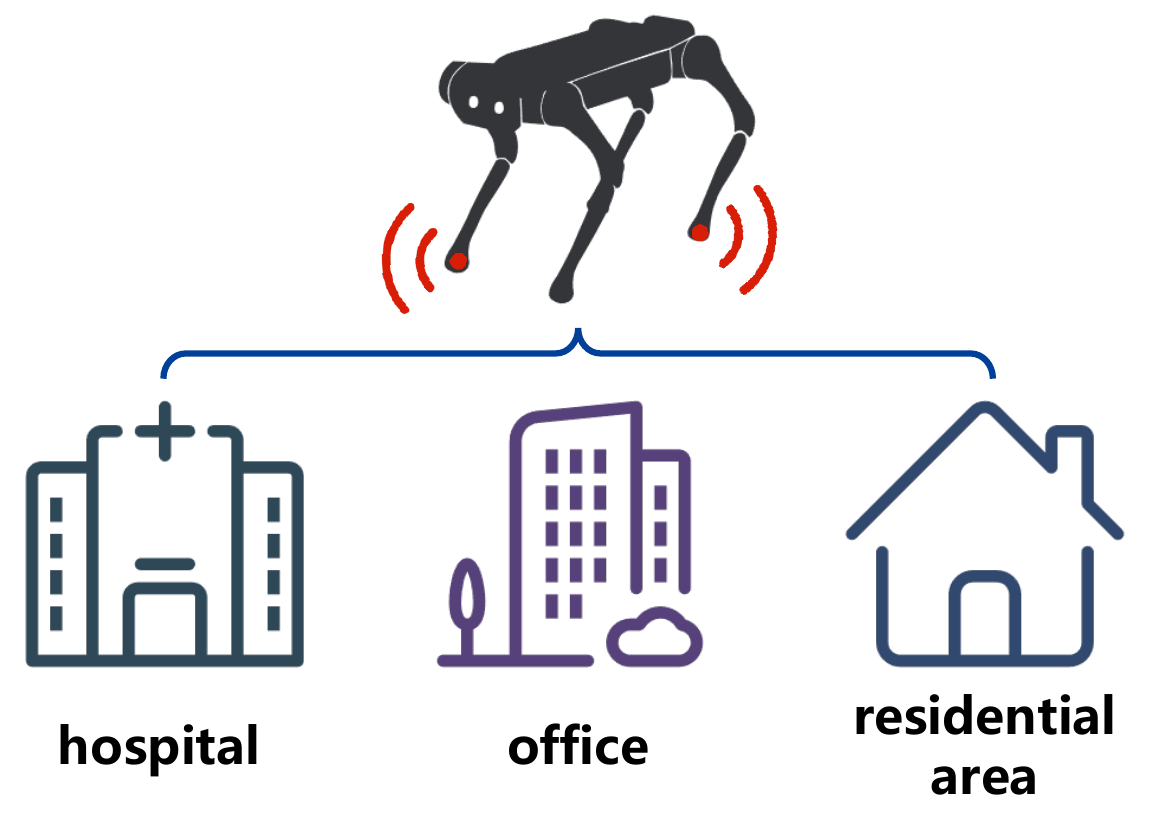}
    \caption{\textbf{Noise-sensitive indoor scenarios.} Quadruped robots operating in noise-sensitive environments, such as hospitals, offices, and residential spaces, must minimize noise levels to maintain a comfortable atmosphere.}
    \label{fig:robot_noise}
\end{figure}

\subsection{Noise in Human-Robot Interaction}
In human-robot interaction (HRI) scenarios, noise generated by robots can significantly impact the user experience. Previous research has investigated the relationship between robot noise characteristics and user perception. For instance, a pilot study conducted by Zhang et al.~\cite{zhang2021exploring} indicates that robots producing quieter and higher-pitched sounds are often perceived as exhibiting greater competence and eliciting less discomfort from users. The findings from their study further reinforce the notion that quieter robots are generally regarded as more comfortable, emphasizing the crucial role of auditory design in robot locomotion for shaping the user experience during interactions. Supporting this concept, Izui et al.~\cite{izui2020correlation} demonstrate that motor noise can directly affect users' impressions of a robot's movement, underscoring the importance of noise management in enhancing human-robot interaction. These studies highlight that quieter robot design contributes to greater comfort during HRI, underscoring the critical role of auditory design in improving user experience and fostering more seamless human-robot interaction.

\subsection{Noise in Quadruped Robots}
The primary source of noise in quadruped robots during locomotion is the repetitive impact of the robot's feet striking the ground~\cite{christie2016acoustics}. These impacts generate considerable noise, particularly on hard surfaces such as tiles or wooden floors, where sound propagation is more pronounced compared to other surfaces. Besides, the robot's motor and joint mechanisms can also contribute to noise generation, particularly when operating at high speeds. The combination of these factors results in a significant acoustic noise, limiting the robot's suitability for noise-sensitive indoor applications.

In indoor environments, the operating speed of robots is generally limited, which results in lower levels of motor noise. While noise generated by motors and joint mechanisms is influenced by the robot's structural design and can be reduced through optimization techniques, such as minimizing mechanical friction~\cite{gonzalez2023noise}, it tends to be relatively minor compared to the noise produced by foot-ground collisions, and thus is not the primary focus of this study. Therefore, this work primarily focuses on optimizing control algorithms to minimize foot-ground impact noise, without altering the robot's hardware design, to enhance its suitability for indoor applications.

Optimizing noise reduction through control algorithms presents a compelling solution, as it allows for the mitigation of impact noise without altering the robot's physical structure. By fine-tuning gait patterns, adjusting the timing and velocity of foot-ground contact, and dynamically controlling limb movements, it is possible to reduce noise output while preserving locomotion stability and efficiency. These control strategies offer the flexibility to adapt to various surfaces and environments, making them highly applicable in noise-sensitive indoor scenarios.

\subsection{Contributions}
In this study, we propose a comprehensive approach to \textbf{M}inimizing aco\textbf{U}s\textbf{T}ic nois\textbf{E} (\textbf{MUTE}) generated by quadruped robots, focusing on the development of advanced motion control algorithms. Our method achieves smoother and quieter movement in the given hardware design, relying solely on optimized control strategies. The effectiveness of our method is systematically assessed through a series of experiments conducted in various indoor environments. The experimental results reveal substantial noise reductions using MUTE compared to previous studies, highlighting the effectiveness of our approach in enhancing the suitability of quadruped robots for noise-sensitive environments.

Our primary contributions are as follows:

\begin{enumerate}
    \item We propose a novel approach, \textbf{MUTE}, to reduce the acoustic noise of quadruped robots through gait optimization and a quiet factor \( \beta \) that balances noise reduction and agility. Refining the robot's motion control algorithms enables significant noise reduction during locomotion, enhancing its suitability for noise-sensitive indoor environments while allowing adaptability across various surfaces.
    \item We establish a comprehensive evaluation framework to assess the robot's noise levels quantitatively in diverse indoor settings. Measurement of the Mean Noise Level (MNL) and Peak Noise Level (PNL) provides an in-depth analysis of the robot's acoustic performance across different surface materials.
    \item We experimentally validate the effectiveness of MUTE, demonstrating an average noise reduction of 8 dBA compared to baseline methods at equivalent locomotion speeds. Additionally, in experiments conducted in  actual office environments, our approach maintains noise levels at a threshold that does not pose any harm to human health.
\end{enumerate}

\section{METHODOLOGY}

The amplitude of sound generated by an impact is generally proportional to the square root of the impact energy~\cite{van1998sounds}. Given foot-ground collisions, the energy loss is associated with the foot's kinetic energy, expressed as \( E = \frac{1}{2}mv^2 \). Thus, the sound amplitude is directly proportional to the foot's velocity at the moment of contact. Given the constraints of acoustic modeling in physical simulations, this study addresses the problem by imposing constraints on the foot's velocity just prior to contact. This approach enables indirect control of noise produced during foot-ground collisions by focusing on minimizing impact velocity.

\subsection{Preliminaries}

Due to the absence of exteroceptive sensors, the problem is modeled as an infinite-horizon Partially Observable Markov Decision Process (POMDP), represented by the tuple \(M = (\mathcal{S}, \mathcal{A}, \mathcal{T}, \mathcal{R}, \Omega, \mathcal{O}, \gamma)\). In this formulation, the agent's policy \(\pi\) selects an action \(a_t \in \mathcal{A}\) based on the current state \(s_t \in \mathcal{S}\), and the environment transitions to the next state \(s_{t+1}\) according to the transition function \(\mathcal{T}(s_{t+1} | s_t, a_t)\). The agent then receives a reward \(r_t\) from the reward function \(\mathcal{R}(s_t, a_t)\) and observes \(o_t \in \Omega\) through the observation function \(\mathcal{O}(o_t | s_{t+1}, a_t)\). The objective is to determine an optimal policy \(\pi^*\) that maximizes the expected cumulative reward \(J(\pi)\) over an infinite horizon, expressed as:
\begin{equation}
    J(\pi) = \mathbb{E}_{\pi} \left[ \sum_{t=0}^{\infty} \gamma^t \mathcal{R}(s_t, a_t) \right],
\end{equation}
where \(\gamma \in [0, 1)\) is the discount factor.

Additionally, the historical sequence of observations, denoted as \( o^H_t \), where \( H \) represents the length of the observation history window, is utilized to capture temporal dependencies. The implicit encoding of this sequence is represented by \( z_t \).

\begin{figure*}[t]
    \centering
    \includegraphics[width=0.8\textwidth]{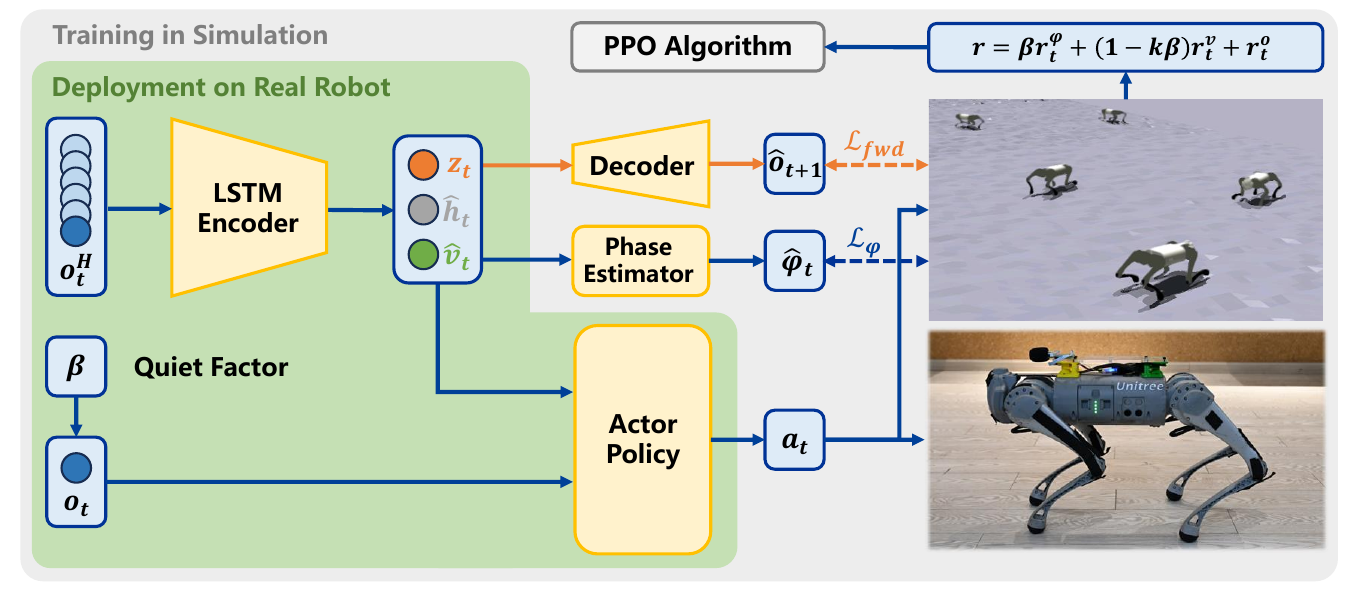}
    \caption{\textbf{Proposed Controller Structure.} The controller features a Long Short-Term Memory (LSTM) encoder-decoder network for phase estimation and latent variable representation. The quiet factor \( \beta \) serves as an input to the policy network, which is trained to balance noise reduction and speed. The gray area indicates the training phase conducted in simulation, while the green area highlights the transition to real-world implementation.}
    \label{fig:controller}
\end{figure*}

\subsection{Phase Estimation}

Accurate estimation of the moments preceding foot-ground contact is crucial for effectively implementing velocity constraints on the foot. To facilitate this, we introduce the concept of phase $\varphi_{t,i}$, where $t$ represents the discrete time index, and $i \in \{1, 2, 3, 4\}$ denotes the leg index. The phase variable $\varphi_{t,i}$ is continuous and takes values within the interval $\varphi_{t,i} \in [0, 1]$, where $\varphi_{t,i} = 0$ denotes the instant of leg lift-off, and $\varphi_{t,i} = 1$ denotes the moment of ground contact. The phase \(\varphi_{t, i}\) transitions uniformly between these two states over time, completing a full cycle that corresponds to the robot's gait cycle. When \(\varphi\) progresses from 0 to 1, it reflects the leg's swing phase, indicating airborne movement. Conversely, the transition from 1 back to 0 represents the stance phase, during which the leg maintains ground contact. This cyclical progression captures the sequential phases of the robot's gait.


To enhance phase prediction accuracy, an LSTM network processes the historical observation sequence \( o^H_t \), enabling estimation of the terrain height map \( h_t \) surrounding the robot. Building on the significance of velocity estimation highlighted by~\cite{wang2024understanding, nahrendra2023dreamwaq}, velocity \( v_t \) is also estimated from the same observation history. Additionally, to capture comprehensive information from the sequence, a forward prediction model is trained to represent the latent variable \( z_t \). This model uses the LSTM-encoded \( z_t \) and employs a decoder to reconstruct the next observation \( o_{t+1} \), following Variational Autoencoder (VAE)~\cite{kingma2013auto, higgins2017beta, burgess2018understanding}. Finally, the estimated \( \hat{h}_t \) and \( \hat{v}_t \), together with the latent variable \( z_t \), are input into a MLP network to estimate the phase \( \hat{\varphi}_t \).

The loss function of the phase estimation network comprises three components: the estimation loss \( \mathcal{L}_{\operatorname{est}} \) for velocity \( v_t \) and terrain height \( h_t \), the forward model loss \( \mathcal{L}_{\operatorname{fwd}} \), and the phase estimation loss \( \mathcal{L}_{\varphi} \). The total loss is defined as:
\begin{equation}
    \mathcal{L} = \mathcal{L}_{\operatorname{est}} + \lambda_{\operatorname{fwd}} \mathcal{L}_{\operatorname{fwd}} + \lambda_{\varphi} \mathcal{L}_{\varphi}.
\end{equation}
where \( \lambda_{\operatorname{fwd}} \) and \( \lambda_{\varphi} \) are the weighting factors that control the relative importance of the forward model loss and phase estimation loss in the total loss function, respectively.

\begin{enumerate}
    \item The estimation loss \( \mathcal{L}_{\operatorname{est}} \) is computed as the Mean Squared Error (MSE) between the estimated and true values of velocity and terrain height map obtained from the simulation environment, expressed as:
    \begin{equation}
        \mathcal{L}_{\operatorname{est}} = \operatorname{MSE}(v_t, \hat{v}_t) + \operatorname{MSE}(h_t, \hat{h}_t).
    \end{equation}

    \item The forward model loss \( \mathcal{L}_{\operatorname{fwd}} \) is derived from the reconstruction loss of the forward model and the Kullback-Leibler (KL) divergence, which regularizes the latent variable, defined as:
    \begin{equation}
        \mathcal{L}_{\operatorname{fwd}} = \operatorname{MSE}(o_{t+1}, \hat{o}_{t+1}) + \lambda D_{KL}(q(z_t | o^H_t) || N(0, I)).
    \end{equation}

    \item The phase estimation loss \( \mathcal{L}_{\varphi} \) is calculated as the MSE between the predicted phase values and the true phase values, formulated as:
    \begin{equation}
        \mathcal{L}_{\varphi} = \operatorname{MSE}(\varphi_t, \hat{\varphi}_t),
    \end{equation}
    where \( \varphi_t \) represents the ground truth phase values derived from the simulation. Specifically, \( \varphi_t \) takes on the values of 0 and 1 at the moments of foot-off and foot contact, respectively, while the phase values at intermediate time steps are obtained through linear interpolation between these two instances.
\end{enumerate}

\subsection{Quiet Factor}

The robot's speed has a direct impact on the noise it generates. In certain scenarios, fast movement is necessary, even at the cost of increased noise, while in others, noise reduction becomes a priority, potentially requiring a reduction in speed. To balance this trade-off between speed and noise, we introduce a quiet factor \( \beta \), a scalar value ranging from 0 to 1. This factor regulates the balance: \( \beta = 0 \) prioritizes speed over noise reduction, while \( \beta = 1 \) emphasizes noise minimization at the cost of speed.

The quiet factor \( \beta \) is integrated into the policy network as part of the command input \( c_t \), alongside the desired velocities \( v_x^{\operatorname{cmd}} \), \( v_y^{\operatorname{cmd}} \), and \( w_z^{\operatorname{cmd}} \). By modulating these velocity commands based on the value of \( \beta \), the robot can dynamically adjust its behavior to achieve the desired balance between noise reduction and locomotion speed.

To facilitate exploration across various noise-speed trade-offs, the quiet factor \( \beta \) is sampled from a uniform distribution at the beginning of each agent reset. This factor directly affects the reward function by scaling both phase-related and task-related rewards according to \( \beta \). The total reward is computed as:
\begin{equation}
    r = \beta r^{\varphi} + (1 - k\beta) r^{v} + r^{o},
\end{equation}
where \( r^{\varphi} \) is the phase-related reward for gait optimization, \( r^{v} \) represents the task-related reward for tracking velocity commands, and \( r^{o} \) encompasses additional reward terms. The parameter \( k \) determines the degree of speed reduction in favor of noise minimization. For the experiments, \( k \) is set to 0.2 to achieve a balanced trade-off between speed and noise reduction.

\subsection{Training Framework}

To address the challenge posed by the partial observability of the environment, the training framework adopts an Asymmetric Actor-Critic (AAC) architecture~\cite{pinto2017asymmetric}. In this structure, the actor network is tasked with predicting the optimal action \( a_t \) based on the current observation \( o_t \), while the critic network evaluates the value function using the full state information \( s_t \). Both networks are trained simultaneously using the Proximal Policy Optimization (PPO) algorithm~\cite{schulman2017proximal}, ensuring stability and efficiency during policy updates.

\subsubsection{Policy Network}

The policy network \( \pi (a_t | o_t, h_t, v_t, z_t) \) generates the optimal action \( a_t \) based on the current observation \( o_t \), the estimated terrain height map \( h_t \), body velocity \( v_t \), and latent variable \( z_t \). The observation \( o_t \) is derived from the robot's proprioceptive sensors and is defined as:
\begin{equation}
    o_t = [\omega_t, g_t, q_t, \dot q_t, a_{t-1}, c_t],
\end{equation}
where \( \omega_t \) denotes the angular velocity, \( g_t \) represents the gravity vector, \( q_t \) and \( \dot q_t \) correspond to the joint angles and velocities, \( a_{t-1} \) is the action taken in the previous timestep, and \( c_t \) represents the command input.

\subsubsection{Action Space}

The output action \( a_t \in \mathbb{R}^{12} \) represents the desired joint position offsets relative to the constant standing joint positions \( q^{\operatorname{stand}} \). Consequently, the final desired joint positions are given by:
\[
    q^{\operatorname{des}}_t = q^{\operatorname{stand}} + a_t.
\]

The motors then track these desired joint positions using a Proportional-Derivative (PD) controller, with gains set to \( K_p = 20 \) and \( K_d = 0.5 \). The PD controller adjusts the joint positions by minimizing the error between the actual and desired joint angles. Specifically, \( K_p \) governs the proportional response to the positional error, while \( K_d \) provides velocity-based damping to smooth the motion.

\subsubsection{Value Network}

The value network \( V(s_t) \) estimates the value function using the full state information \( s_t \), defined as:
\begin{equation}
    s_t = [o_t, \hat{h}_t, \hat{v}_t, z_t, o^{\operatorname{priv}}_t],
\end{equation}
where \( o^{\operatorname{priv}}_t \) represents the privileged information, including the true terrain height map, body velocity, foot contact information, and phase ground truth. The value function is used to evaluate the quality of the current state and guide policy updates during training. By incorporating the full state information, the value network provides a comprehensive assessment of the state's value, enabling more informed policy decisions.

\begin{table}[h]
    \renewcommand{\arraystretch}{1.2}
    \caption{Reward terms}
    \label{table:reward}
    \begin{center}
        \scalebox{0.9}{
            \begin{tabular}{c|c|c|c}
                \hline
                \textbf{Term}               & \textbf{Reward}        & \textbf{Equation}                                 & \textbf{Weight} \\
                \hline
                \multirow{2}{*}{Task}       & Lin. velocity tracking & \( \exp\{-4(v_{xy}^{\operatorname{cmd}} - v_{xy})^2\} \)         & 1.0             \\
                                            & Ang. velocity tracking & \( \exp\{-4(\omega_{z}^{\operatorname{cmd}} - \omega_{z})^2\} \) & 0.5             \\
                \hline
                \multirow{2}{*}{Phase}      & Drop Foot velocity     & \( e^{ \hat{\varphi}} v_{f, \operatorname{drop}}^2 \)            & -0.05           \\
                                            & Raise Foot velocity    & \( e^{ -\hat{\varphi}} v_{f, \operatorname{raise}}^2 \)          & 0.01            \\
                \hline
                \multirow{5}{*}{Smoothness} & Linear velocity (z)    & \( v_z^2 \)                                       & -2.0            \\
                                            & Angular velocity (xy)  & \( \omega_{xy}^2 \)                               & -0.05           \\
                                            & Joint power            & \( |\tau||\dot{q}| \)                             & -2.0e-5         \\
                                            & Joint acceleration     & \( \Vert \ddot{q} \Vert^2 \)                      & -2.5e-7         \\
                                            & Action rate            & \( \Vert a_t - a_{t-1} \Vert^2 \)                 & -0.01           \\
                \hline
                \multirow{1}{*}{Safety}     & Collision              & \( n_{\operatorname{collision}}\)                                & -0.1            \\

                \hline
                \multirow{2}{*}{Body}       & Body height            & \( \Vert h^{\operatorname{des}} - h \Vert ^2 \)                  & -1.0            \\
                                            & Orientation (gravity)  & \( \Vert g \Vert^2 \)                             & -0.2            \\
                \hline
            \end{tabular}
        }
    \end{center}
\end{table}

\subsubsection{Reward Terms Design}

The reward function \( \mathcal{R}(s_t, a_t) \) is crafted to promote noise reduction while ensuring locomotion stability and efficiency. This reward signal comprises multiple components, each targeting specific aspects of the robot's behavior. The components include task-related rewards, phase-related rewards, smoothness penalties, safety considerations, and body orientation constraints. A summary of these reward terms and their corresponding equations is presented in Table~\ref{table:reward}.

In designing the phase-related rewards, the estimated phase \( \hat{\varphi}_t \) serves as the basis. Penalty functions are applied to drop foot velocities, with penalties increasing as the phase nears 1. Additionally, a minor reward is assigned to raise foot velocities to encourage consistent foot elevation within each cycle. This reward is more significant when the phase is closer to 0. The phase-related reward term is thus defined as follows:
\begin{equation}
    r^{\varphi} = w_{d} \sum_{i=1}^{4} e^{\hat{\varphi}_i} v_{f_i, \operatorname{drop}}^2 + w_{r} e^{-\hat{\varphi}_i} \sum_{i=1}^{4} v_{f_i, \operatorname{raise}}^2,
\end{equation}
where \( v_{f_i, \operatorname{drop}} \) and \( v_{f_i, \operatorname{raise}} \) represent the drop and raise foot velocities of the \( i \)-th foot, respectively. The associated weights are denoted by \( w_{d} \) and \( w_{r} \).

\section{EXPERIMENTAL RESULTS}

To evaluate the effectiveness of the proposed \textbf{MUTE} approach in reducing noise generated by quadruped robots during locomotion, a series of experiments are conducted in real-world settings. These experiments aim to address the following research questions:

\begin{enumerate}
    \item Can the MUTE approach effectively reduce noise at identical speed commands?
    \item Does MUTE maintain its effectiveness in noise reduction when the robot operates at the same actual speed?
    \item How does the Quiet Factor \( \beta \) influence the balance between speed and noise?
    \item Can MUTE sustain a low noise level during prolonged usage in real-world scenarios?
\end{enumerate}

\subsection{Experimental Setup}
\subsubsection{Simulation Training}

We use the Unitree Go1 EDU quadruped robot as the agent for training. The Proximal Policy Optimization (PPO) algorithm~\cite{schulman2017proximal} is employed, with policy optimization conducted via the Adam optimizer at a learning rate of \( 10^{-3} \). The training utilizes 4096 parallel agents within Isaac Gym~\cite{rudin2022learning, makoviychuk2021isaac}, running for a total of 6000 episodes. This process takes approximately 8 hours to complete on a single NVIDIA GeForce RTX 3090Ti GPU.

To ensure the development of robust locomotion policies, curriculum learning is applied, gradually increasing terrain complexity over five levels. The robots progressively learn locomotion strategies on a variety of terrains, including flat surfaces, rough terrains, discretized terrains, and stairs. This approach facilitates the agent's ability to generalize across diverse environments, enabling effective adaptation to challenging real-world scenarios.

\subsubsection{Real-World Transfer}

Upon completion of the simulation training, the learned policy is transferred to the real-world Unitree Go1 EDU robot without further tuning. The robot relies solely on proprioceptive sensors, including joint angles, velocities, and body orientation, without external sensor input. The control policy is executed at a frequency of 50 Hz on a personal computer equipped with an Intel Core i9-12900H CPU.

\begin{figure}[htbp]
    \centering
    \subfloat[Simulation]{\includegraphics[width=0.22\textwidth]{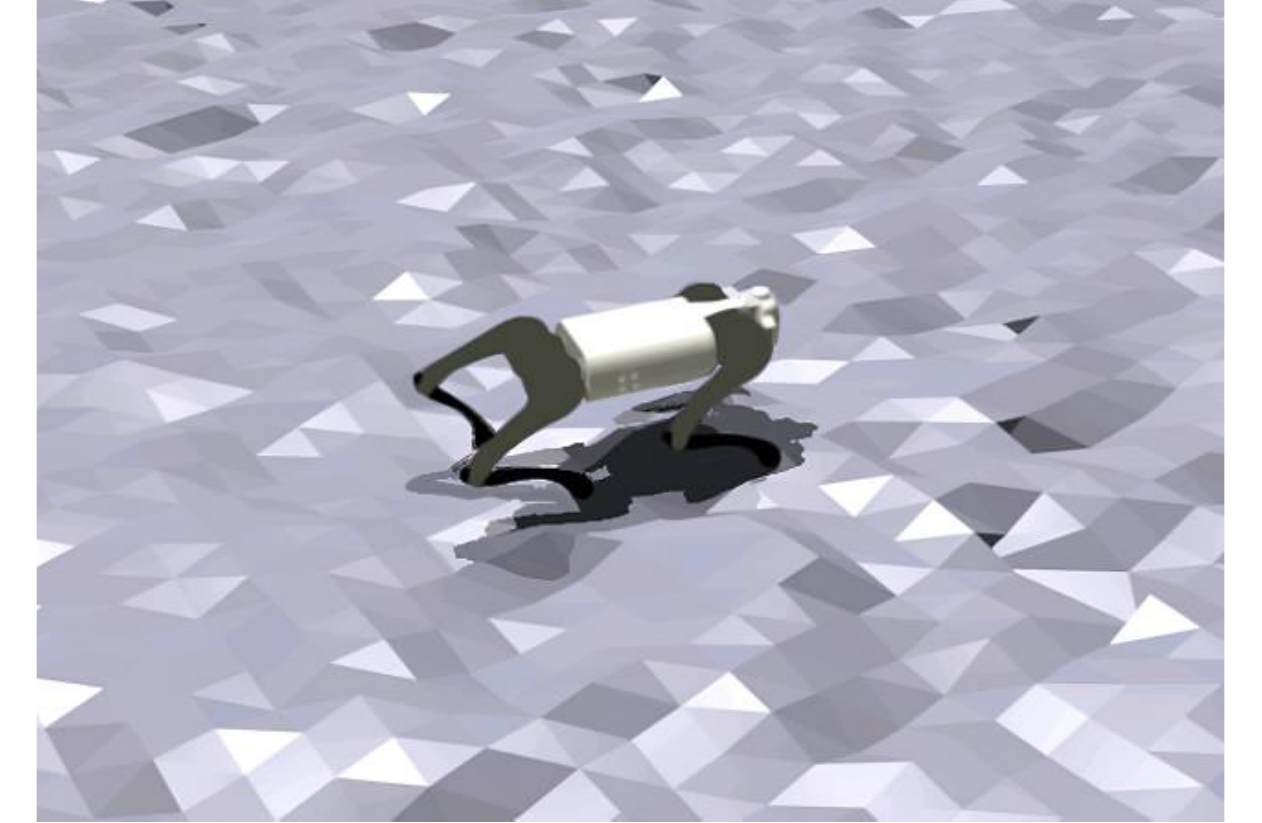}\label{fig:sim}}
    \hfill
    \subfloat[Real robot]{\includegraphics[width=0.22\textwidth]{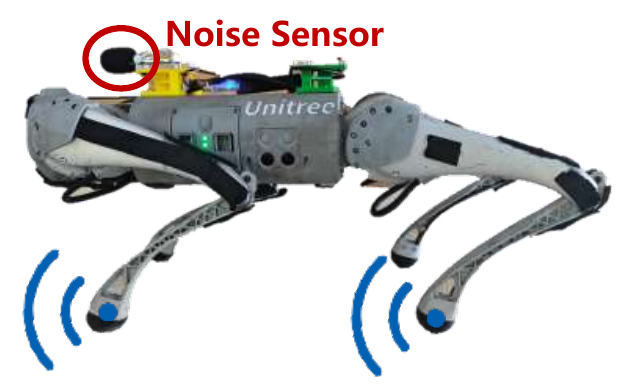}\label{fig:real}}
    \caption{\textbf{Experimental Setup for Simulation and Real-World Transfer.} The left panel depicts the simulation environment used for training in Isaac Gym, while the right panel presents the real robot setup, equipped with a sound pressure sensor for noise measurement in actual environments. The sensor updates at a frequency of 20 Hz to capture real-time noise levels during robot locomotion.}
    \label{fig:sim_real}
\end{figure}

As illustrated in Fig.~\ref{fig:real}, we measure the noise level during robot locomotion utilizing a sound pressure sensor mounted approximately 30 cm above the ground. The sensor records sound pressure levels in decibels (dBA) while the robot is in motion.


\subsection{Evaluation Metrics}

To assess the noise produced by the robot during locomotion, sound pressure levels are measured in decibels (dBA) using a noise sensor mounted on the robot. The A-weighted decibel scale is selected due to its better alignment with human auditory sensitivity across different frequencies, thereby providing a more reasonable estimate of perceived noise levels~\cite{international2013electroacoustics}.

Inspired by \cite{fletcher1933loudness}, the sound pressure level (SPL) in dBA is calculated using the following formula:
\begin{equation}
    \text{SPL (dBA)} = 20 \log_{10} \left( \frac{p}{p_0} \right),
\end{equation}
where \( p \) represents the root mean square sound pressure and \( p_0 \) is the reference sound pressure, typically \( 20 \mu\text{Pa} \) in air.

We define two key metrics for evaluating the robot's acoustic performance:
\begin{itemize}
    \item \textbf{Mean Noise Level (MNL)}: This metric represents the average sound pressure level recorded during the robot's locomotion, providing an overall measure of the typical noise generated during movement. MNL is particularly useful for assessing the sustained or continuous impact of noise on human auditory perception, offering an indication of the robot's environmental noise footprint over time.
    \item \textbf{Peak Noise Level (PNL)}: This metric captures the maximum sound pressure level observed during locomotion, highlighting the loudest noise events, which significantly influence human auditory perception. PNL is critical for understanding the instantaneous noise impact, particularly in scenarios where transient noise spikes may cause discomfort or alarm.
\end{itemize}

Measurement of both the MNL and PNL provides a comprehensive understanding of the robot's noise characteristics across different velocity commands and surface types.

\subsection{Compared Methods}

To evaluate the effectiveness of MUTE in reducing noise, a comparison is conducted against two baseline locomotion policies. The methods included in this comparison are:
\begin{enumerate}
    \item \textbf{DreamWaQ}~\cite{nahrendra2023dreamwaq}: A state-of-the-art locomotion policy optimized for efficient terrain traversal and stability, trained using the DreamWaQ algorithm.
    \item \textbf{Built-in MPC}: The default Model Predictive Control (MPC) policy integrated into the Unitree Go1 EDU robot, optimized for fundamental locomotion tasks.
    \item \textbf{MUTE w/ \( \beta = 0 \)}: The proposed method with the quiet factor \( \beta \) set to 0, prioritizing speed without active noise reduction.
    \item \textbf{MUTE w/ \( \beta = 1 \)}: The proposed method with the quiet factor \( \beta \) set to 1, emphasizing noise minimization over speed.
\end{enumerate}

These baseline methods vary in complexity and adaptability in locomotion. By comparing the noise levels generated by MUTE with these baselines, the effectiveness in reducing the robot's overall acoustic noise while maintaining locomotion can be assessed.


\subsection{Noise Reduction Performance}

The noise generated by the robot during locomotion is measured on three common indoor surfaces: wooden flooring, carpet, and porcelain tiles. For each surface, the robot is instructed to traverse at a constant speed of 0.5 m/s for at least 10 seconds, with five trials conducted per surface. Both the MNL and PNL are recorded for each trial, and the results are presented in Table~\ref{table:noise}. When the robot is stationary, with only the motor fan running, the noise level is approximately 55 dBA.

\begin{table}[htbp]
    \renewcommand{\arraystretch}{1.2}
    \caption{Noise levels on different surfaces at 0.5 m/s}
    \label{table:noise}
    \begin{center}
        \scalebox{1.0}{
            \begin{tabular}{c|c|c|c}
                \hline
                \textbf{Method}                                  & \textbf{Surface} & \textbf{MNL (dBA)}          & \textbf{PNL (dBA)}          \\
                \hline
                \multirow{4}{*}{DreamWaQ}                        & Wood             & \(73.51 \pm 1.44\)          & \(83.18 \pm 1.16\)          \\
                                                                 & Carpet           & \(71.90 \pm 0.47\)          & \(80.80 \pm 1.82\)          \\
                                                                 & Tiles            & \(72.34 \pm 1.31\)          & \(80.93 \pm 1.80\)          \\
                                                                 \cline{2-4}
                                                                 & \textbf{Average} & \(72.58\)          & \(81.64\)          \\
                \hline
                \multirow{4}{*}{Built-in MPC}                    & Wood             & \(79.74 \pm 1.04\)          & \(84.03 \pm 1.03\)          \\
                                                                 & Carpet           & \(77.92 \pm 1.21\)          & \(81.25 \pm 1.29\)          \\
                                                                 & Tiles            & \(79.38 \pm 1.91\)          & \(83.12 \pm 1.08\)          \\
                                                                 \cline{2-4}
                                                                 & \textbf{Average} & \(79.01\)          & \(82.80\)          \\
                \hline
                \multirow{4}{*}{\textbf{MUTE w/ \( \beta = 0\)}} & Wood             & \(68.61 \pm 0.79\)          & \(75.86 \pm 1.70\)          \\
                                                                 & Carpet           & \(67.42 \pm 0.73\)          & \(74.40 \pm 2.91\)          \\
                                                                 & Tiles            & \(70.69 \pm 0.53\)          & \(79.30 \pm 1.20\)          \\
                                                                 \cline{2-4}
                                                                 & \textbf{Average} & \(68.91\)          & \(76.52\)          \\
                \hline
                \multirow{4}{*}{\textbf{MUTE w/ \( \beta = 1\)}} & Wood             & \(\textbf{65.78} \pm 0.87\) & \(\textbf{73.18} \pm 1.83\) \\
                                                                 & Carpet           & \(\textbf{62.48} \pm 0.58\) & \(\textbf{69.52} \pm 1.02\) \\
                                                                 & Tiles            & \(\textbf{66.15} \pm 0.66\) & \(\textbf{76.22} \pm 1.58\) \\
                                                                 \cline{2-4}
                                                                 & \textbf{Average} & \(\textbf{64.80}\) & \(\textbf{72.97}\) \\
                \hline
            \end{tabular}
        }
    \end{center}
\end{table}

The results in Table~\ref{table:noise} demonstrate that MUTE consistently outperforms the baseline policies in reducing noise levels across all tested surfaces. When the quiet factor \( \beta \) is set to 1, the robot achieves the lowest noise levels on all surfaces, with active velocity concessions. Notably, when \( \beta \) is set to 0, MUTE still maintains lower noise levels compared to the baseline policies, underscoring its effectiveness in noise reduction while ensuring locomotion performance. Additionally, on surfaces particularly prone to noise, such as wood and tiles, MUTE exhibits the most significant reductions, highlighting its potential for applications in noise-sensitive indoor environments.

\begin{figure}[htbp]
    \centering
    \includegraphics[width=0.45\textwidth]{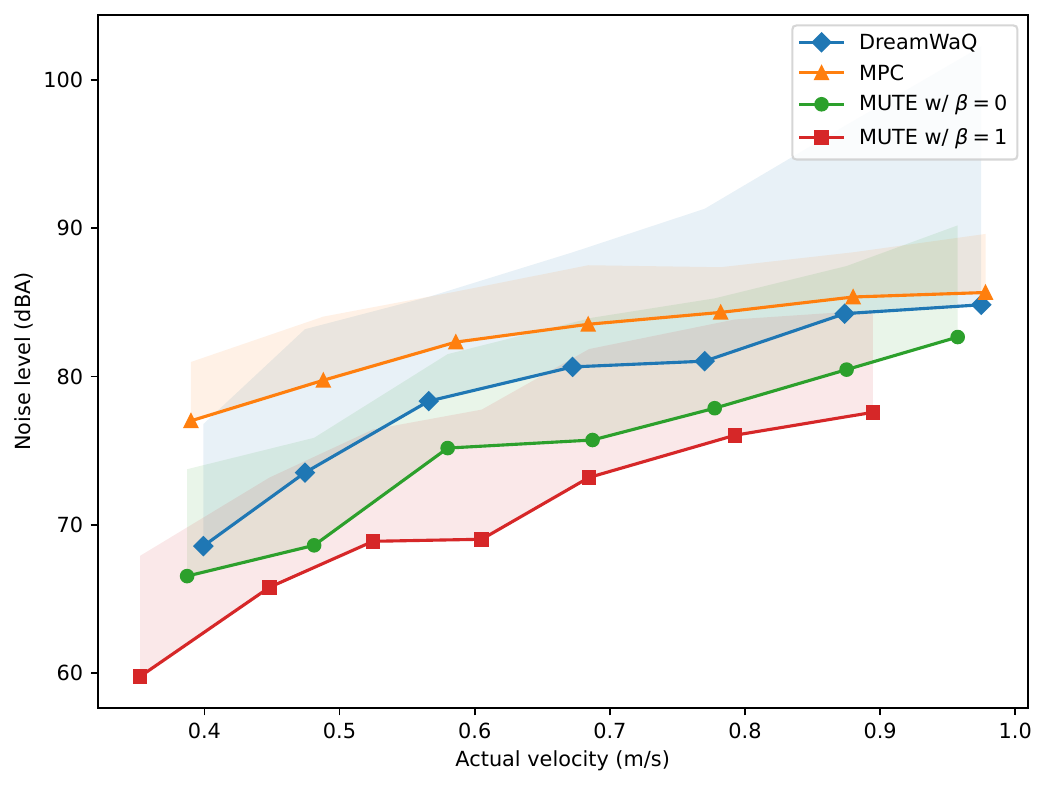}
    \caption{\textbf{Noise levels on wooden flooring at different actual speeds.} The solid lines represent the MNL, while the light-colored bands indicate the PNL. The $x$-axis denotes the robot's actual speed, calculated using EKF~\cite{bloesch2013state, bledt2018mit}, rather than the commanded speed. MUTE consistently achieves the lowest noise levels across all speed conditions, with the quiet factor \( \beta = 1 \) providing the most significant noise reduction.}
    \label{fig:wood}
\end{figure}

To further explore the relationship between noise levels and locomotion speed, experiments are conducted on wooden flooring with speed commands ranging from 0.4 m/s to 1.0 m/s, repeating each trial five times. As shown in Fig.~\ref{fig:wood}, noise levels increase with the robot's actual speed. MUTE with \( \beta = 1 \) consistently achieves the lowest MNL and PNL across all speed conditions, demonstrating its effectiveness in noise reduction while maintaining comparable speeds to the baseline policies. On average, MUTE reduces noise levels by approximately 8 dBA compared to the baseline, equating to a 2.5-fold decrease in sound pressure. Additionally, even with \( \beta = 0 \), MUTE produces lower noise levels than the baseline policies, underscoring its ability to reduce noise while preserving adaptable locomotion performance.

\begin{figure*}
    \centering
    \includegraphics[width=0.85\textwidth]{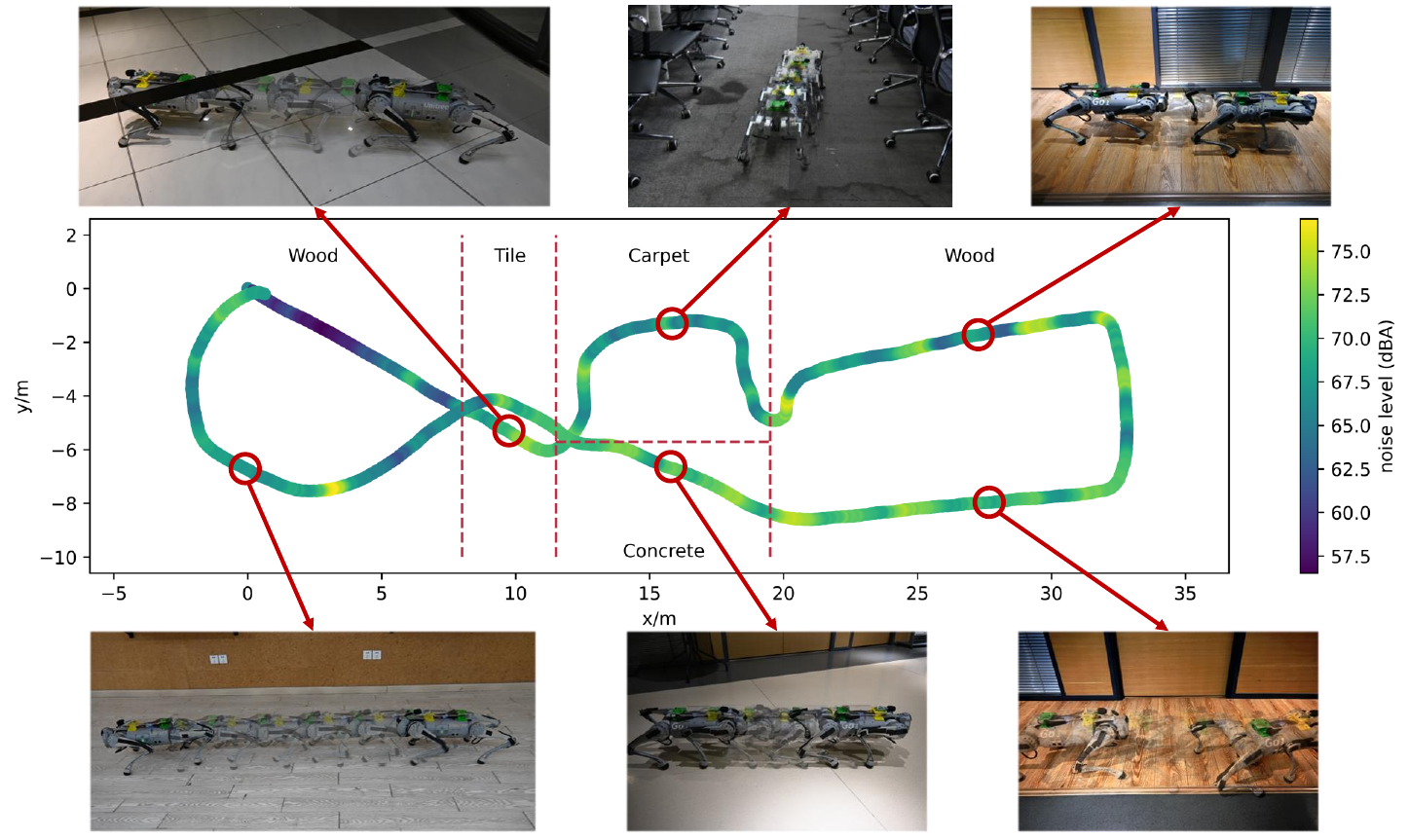}
    \caption{\textbf{Noise levels during long-distance walking on mixed surfaces.} The colored line represents the robot's path, with color variations indicating noise levels at different points. Dotted lines divide the various surface types. The total path is approximately 91.7 meters, with the robot maintaining an average speed of 0.36 m/s, resulting in an MNL of 68.25 dBA and a PNL of 76.8 dBA.}
    \label{fig:long_distance}
\end{figure*}

\subsection{Quiet Factor Analysis}

To evaluate the impact of the quiet factor \( \beta \) on noise levels, experiments are conducted with \( \beta \) ranging from 0 to 1. The robot traverses wooden flooring at a constant speed of 0.8 m/s, with each trial repeated five times. For every trial, noise levels and velocity tracking squared errors are recorded to analyze the trade-off between noise reduction and speed.

\begin{figure}[htbp]
    \centering
    \includegraphics[width=0.45\textwidth]{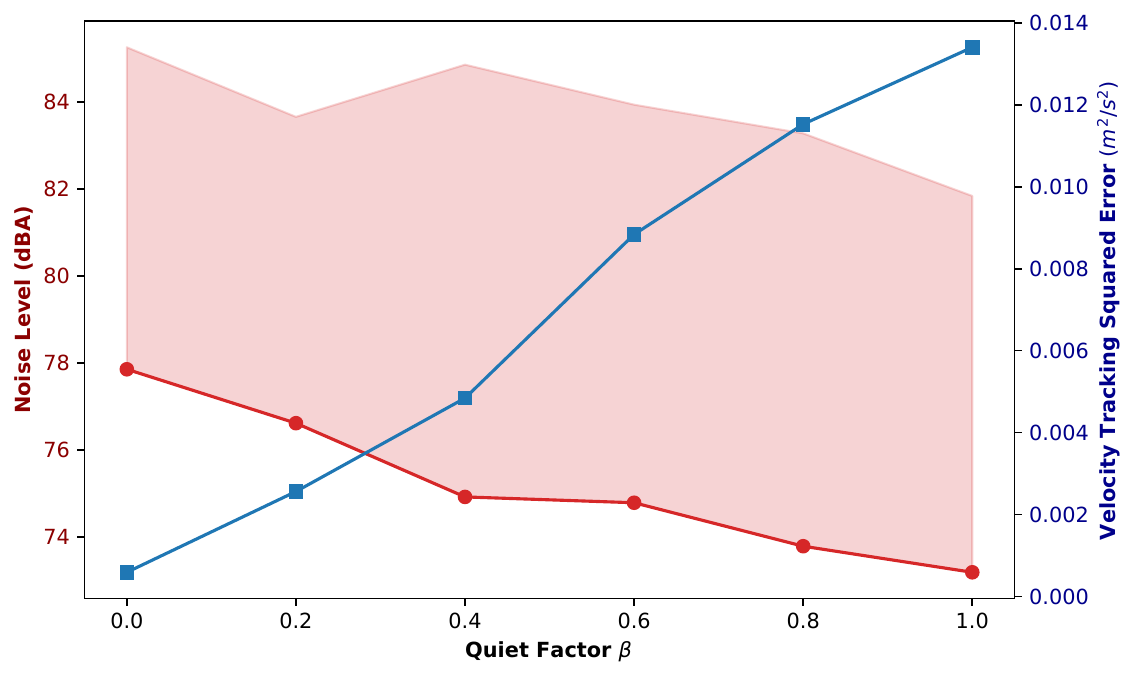}
    \caption{\textbf{Noise levels on wooden flooring at 0.8 m/s command for different \( \beta \) values.} The red line represents the MNL, the light-colored bands indicate the PNL, and the blue line denotes the velocity tracking squared error. Noise levels decrease as \( \beta \) increases, with the MNL dropping to lowest at \( \beta = 1 \). The velocity tracking error shows a increase as \( \beta \) approaches 1, indicating an active compromise in speed.}
    \label{fig:wood_factor}
\end{figure}

Increasing the quiet factor \( \beta \) results in a clear reduction in noise levels, as shown in Fig.~\ref{fig:wood_factor}. At \( \beta = 1 \), which prioritizes noise reduction, the MNL drops to approximately 73.18 dBA, the lowest observed level, while the PNL also remains consistently lower than at other settings. This indicates that focusing on minimizing noise is effective across various test runs. Conversely, at \( \beta = 0 \), where speed is prioritized, noise levels peaked near 77.85 dBA, showing that noisier operation corresponds to a higher performance emphasis on velocity. Additionally, the velocity tracking squared error shows a increase as \( \beta\) approaches 1, suggesting a minor compromise in locomotion speed. This balance is effectively regulated by \( \beta\), enabling the robot to adapt its behavior to various noise-sensitive environments.

These results highlight the flexibility offered by the quiet factor \( \beta \), which allows the robot to adjust its operation to meet the needs of various noise-sensitive environments. In spaces such as hospitals, libraries, and residential areas, where quieter operation is critical, higher \( \beta \) values can ensure a significant reduction in noise levels. Simultaneously, the robot can maintain adequate performance, making it adaptable and functional in diverse real-world scenarios.

\subsection{Long-Distance Walking}

To further evaluate the noise generated during long-distance walking, experiments are conducted in indoor environments featuring mixed surfaces, including wooden flooring, carpet, concrete, and tiles. The robot follows a predefined path through sections of wooden board, ceramic tile, carpet, concrete, and back to the wooden board. The total distance covered is approximately 91.7 meters, with the robot maintaining an average speed of 0.36 m/s, completing the course in 256 seconds.

The MNL recorded during the walk is 68.25 dBA, which falls below the 70 dBA threshold considered safe for human hearing, posing no significant risk of hearing damage~\cite{fink2017safe}. \textbf{Since the noise is measured at the robot's trunk, closer to the foot-ground collision point (the primary sound source), the noise level perceived by humans would actually be lower.} According to the sound attenuation law in air, noise decreases by 6 dB for every doubling of the distance from the source. Given that the measurement point is approximately 30 cm above the ground and about 50 cm from the foot-ground collision point, the actual noise perceived by humans would be significantly lower than the recorded value, remaining well within the acceptable range for indoor environments. This is further corroborated by the supplementary video.

Fig.~\ref{fig:long_distance} shows the noise levels recorded during the long-distance test. To minimize potential interference from ambient noise, the experiment is conducted late at night. These results highlight the robot's ability to maintain low noise levels across various surfaces, demonstrating its suitability for quiet operation in noise-sensitive indoor environments.

\section{CONCLUSIONS AND DISCUSSION}

This study introduces a novel approach to minimizing the acoustic noise of quadruped robots in indoor environments, which is crucial for their deployment in noise-sensitive settings such as hospitals, offices, and residential areas. By optimizing gait design and control strategies, locomotion noise is significantly reduced without notably compromising performance. The method, driven by the quiet factor \( \beta \), effectively balances noise reduction with locomotion efficiency, adapting to various environmental needs. Experiments on different indoor surfaces demonstrate consistent noise reductions, with an average decrease of 8 dBA compared to baseline policies, representing a 2.5-fold drop in sound pressure. Although higher \( \beta \) values result in a increase in velocity tracking error, the impact remains minimal, ensuring effective locomotion.These results underscore the potential of quadruped robots to operate quietly in indoor environments, thereby expanding their practical use in spaces requiring noise control.


\addtolength{\textheight}{-12cm}   









\bibliographystyle{IEEEtran}
\bibliography{IEEEabrv, references}

\end{document}